\documentclass[runningheads]{llncs}

\usepackage{graphicx}
\usepackage{booktabs}
\usepackage{bm}
\usepackage{amsmath}
\usepackage{amssymb}
\usepackage{cleveref}
\usepackage{subcaption}

\begin{document}

\title{Simpler and Stronger Baselines for All-in-One Image Restoration}

\author{Hongyu Du\inst{1} \and
Yuqi Wang\inst{1} \and
Wenlong Jiao\inst{2}\thanks{Corresponding author.} \and
Wenwen Li\inst{3} \and
DongWei Ren\inst{1}}
%
\authorrunning{H. Du et al.} 

\institute{School of Artificial Intelligence, Tianjin University, Tianjin, China\\
\email{\{duhongyu, 3022207327, rendw\}@tju.edu.cn} \and
School of Mathematics, Tianjin University, Tianjin, China\\
\email{wenlong@tju.edu.cn} \and
Stomatological Hospital, Tianjin Medical University, Tianjin 300070, China\\
\email{bjkqyylww@163.com}}
%

\maketitle

\begin{abstract}
Current methods for all-in-one image restoration increasingly rely on complex mechanisms, such as Mixture-of-Experts or prompt learning, to handle diverse degradations.In this work, we challenge this trend by showing that structural symmetry provides a strong and effective inductive bias, which is often sufficient to support high-quality restoration in many practical settings. We propose SymUNet, a minimalist baseline that enforces strict encoder-decoder alignment via simple additive fusion, which outperforms complex state-of-the-art models on standard benchmarks while significantly reducing computational costs. Furthermore, to address the negative transfer observed in highly diverse degradation scenarios (e.g., 5-task settings), we introduce SE-SymUNet, which leverages frozen semantic priors (CLIP) to disentangle conflicting restoration tasks. Extensive experiments demonstrate that our approach establishes a simpler, stronger, and more efficient foundation for multi-degradation learning.

\end{abstract}

\section{Introduction}
\label{sec:intro}

All-in-one image restoration aims to recover high-quality images from input images degraded by multiple factors, such as noise, haze, rain, blur and low-light conditions, using a single model. This paradigm offers significant advantages in real-world applications \cite{jiang2025survey,tian2025degradation,li2022all}, including autonomous driving, surveillance systems and mobile photography, where diverse and often composite degradations should be handled efficiently without task-specific architectures.

Despite progress, recent methods have drifted toward increasing complexity.
Prompt-based models \cite{liao2025prompt,potlapalli2023promptir,liu2025up,kong2024towards,yan2025textual,jiang2024autodir,ai2024multimodal,radford2021learning} rely on multimodal cues to condition degradation-specific behavior. 
Mixture-of-Experts (MoE) architectures \cite{zamfir2025complexity,yang2024language,lin2024unirestorer,zhang2024efficient} deploy specialized sub-networks for distinct degradation types. 
Diffusion-based frameworks \cite{ai2024multimodal,liu2025up,luo2025visual,lin2024unirestorer} leverage generative priors for generalization. 
Recently, agent-based systems \cite{chen2024restoreagent,agenticir} use vision-language models (VLMs) to infer and adapt to degradation types dynamically. 
While these approaches push performance boundaries, their intricate designs (e.g., dynamic routing, large pretrained backbones) inflate computational cost and hinder deployment. Moreover, external priors and complex fusion mechanisms often disturb rather than harness the intrinsic degradation cues already present in the features extracted from degraded images.

In this work, we suggest that well-designed encoder feature extraction inherently embeds degradation-specific information, enabling effective restoration without external complexities. However, common baselines employing asymmetric U-Net architectures, e.g., those derived from Transformer-based models like Restormer \cite{zamir2022restormer}, perform well in single-task restoration but fail in all-in-one scenarios. 
Semantic hierarchy misalignment is the main reason, where the encoder captures multi-level features from low-level degradation details to high-level semantics, but asymmetric designs dilute these cues in the heavier decoder. 
Moreover, due to multi-degradation interference, conflicting signals from heterogeneous degradations clash in unaligned feature spaces, leading to training instabilities that are mitigated in single-task settings by degradation consistency, but are amplified in all-in-one contexts, resulting in inferior generalization across different tasks.

To address these challenges, we suggest a return to simplicity: a symmetric U-Net architecture that aligns feature scales and streamlines cross-scale propagation. This symmetry ensures that degradation-aware features from encoder are preserved without dilution in decoder, rendering simple additive fusion in skip connections sufficient for superior performance. Motivated by this insight, we introduce two strong baselines, SymUNet and SE-SymUnet, for all-in-one image restoration.  
SymUNet is a standard symmetric U-Net that unleashes degradation-aware features through strict encoder-decoder hierarchy alignment, achieving better performance with fewer parameters. 
SE-SymUNet is a semantic-enhanced extension that integrates simple cross-attention with frozen CLIP features to inject high-level degradation priors.
SE-SymUNet yields modest yet consistent gains, validating the robustness of our symmetric core while avoiding the complexity of full VLM integration.

Extensive evaluations on three-task (denoising, dehazing, deraining) and five-task (denoising, dehazing, deraining, deblurring, low-light enhancement) benchmarks show that SymUNet sets new state-of-the-art performance, outperforming complex competitors with reduced computational cost. SE-SymUNet delivers modest yet consistent further gains, validating the effectiveness of both our core symmetric design and simple semantic integration.
In addition, both SymUNet and SE-SymUNet are highly versatile and can be easily extended with advanced modules to make further performance improvements.

The contributions of this work can be summarized as:
\begin{itemize}
\item We reveal that asymmetric architectures are the core bottleneck for preserving degradation cues, driving a paradigm shift toward simple designs for all-in-one image restoration, without complex prompts, dynamic routing, or generative priors.

\item We propose two strong baselines SymUNet and SE-SymUNet that are easier to extend and more computationally feasible than alternatives, filling a critical gap in the field.

\item We conduct extensive experiments and analysis, which enable superior performance on benchmarks and provide insights for future all-in-one restoration architecture development.

\end{itemize}
\section{Related Work}
\label{sec:related_work}

\subsection{Task-Specific Image Restoration}
Traditional image restoration networks were typically designed for individual tasks, such as denoising~\cite{shen2023adaptive,liang2021swinir,Li2025ImageDenoising}, deblurring~\cite{Zhang2026TransformerDeblurring}, dehazing~\cite{song2023vision,cai2016dehazenet}, or deraining~\cite{chen2023learning,xiao2022image,Jiang2024DynamicAssociation}, and each task required a distinct network architecture. Although these models achieved strong results on their respective tasks, deploying multiple networks for different degradations is inefficient and limits practical applicability.
Later, unified architectures~\cite{zamir2022restormer,wang2022uformer,chen2022simple,zamir2021multi,cui2023image,guo2024mambair} enabled a single network to handle multiple restoration tasks without redesigning the architecture for each one. Many of these multi-task networks, such as Restormer~\cite{zamir2022restormer}, adopt skip connections using feature concatenation to preserve detailed encoder features.Although these models perform well on individual tasks, their performance often degrades in multi-task scenarios due to large differences in data distributions and conflicts between tasks. Such conflicts increase the complexity of feature representations, making it harder for the network to converge and learn effective restoration mappings across diverse degradations.
Consequently, both early single-task networks and later unified models struggle to generalize effectively when faced with heterogeneous or composite degradations, motivating the development of all-in-one image restoration approaches that can efficiently handle multiple degradations within a single framework.

\begin{figure*}[t]
    \centering
    \setlength{\abovecaptionskip}{2pt} 
    \setlength{\belowcaptionskip}{0pt}
    \includegraphics[width=1\textwidth]{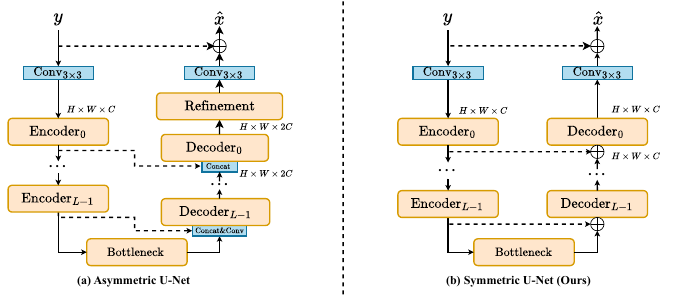}
    \caption{An overview of the architectural philosophies under comparison.
    \textbf{(a) The Asymmetric U-Net} follows a decoder-heavy design widely used in Restormer \cite{zamir2022restormer}, PromptIR \cite{potlapalli2023promptir}, and DFPIR\cite{tian2025degradation}. It adopts concatenation-based skip connections, leading to channel expansion and additional refinement blocks, resulting in a complex and indirect feature fusion path.
    \textbf{(b) Our Symmetric U-Net (SymUNet)} enforces a strict U-shaped design with shared structures across the encoder, decoder, and bottleneck. By using additive skip connections, it maintains consistent channel dimensions and enables a more direct and efficient flow of degradation-aware information.
    }
    \label{fig:architecture}
\end{figure*}

\subsection{All-in-One Image Restoration}
Early all-in-one image restoration work, such as AirNet~\cite{li2022all}, introduced unified CNN-based architectures and leveraged contrastive learning to distinguish degradation types in feature space.
More recent approaches explore prompt-based modalities~\cite{liu2025up,luo2025visual,lin2024improving,ai2024multimodal}, where the restoration network is conditioned on additional task-specific information.
In this framework, visual prompts encode degradation-specific cues through learnable tokens~\cite{potlapalli2023promptir,liu2025uhd,wu2025learning}, while textual or multimodal prompts provide guidance via language or combined visual-language instructions~\cite{yan2025textual,zhang2025perceive,liu2023unifying,tian2024instruct}. Later studies further incorporate pretrained model priors to offer high-level semantic or structural guidance, enhancing generalization~\cite{luo2023controlling,conde2024instructir,jiang2024autodir,yang2024language,chen2025unirestore}.
Other approaches include Mixture-of-Experts (MoE) frameworks~\cite{zhang2024efficient,zamfir2025complexity}, which allocate specialized branches for different degradations, and agent-based systems~\cite{chen2024restoreagent,agenticir}, where vision-language agents reason about degradation types and guide adaptive restoration strategies.
Although these methods improve generalization and adaptability, they rely heavily on external prior knowledge, which can obscure intrinsic degradation features and impede the network from fully learning effective restoration capabilities.
In contrast, a well-designed Transformer backbone~\cite{zamir2022restormer,chen2022simple} can capture both global dependencies and local structures without auxiliary priors, focusing directly on the information present in the degraded image itself.
Motivated by this, we propose a simple Transformer baseline that achieves efficient and effective all-in-one restoration by focusing purely on essential visual information.
\section{Proposed Method}
\label{sec:method}

\subsection{Motivation: Rethinking the Inductive Bias of U-Net}

Significant progress in all-in-one image restoration has largely relied on complex auxiliary modules to adapt standard U-Nets to diverse degradations, while the rationality of the underlying backbone architecture remains underexplored. We observe that most state-of-the-art methods adopt an asymmetric design, which may create a misalignment between encoding and decoding feature manifolds—a sub-optimal inductive bias for multi-degradation learning.

To this end, we shift our focus back to architectural fundamentals and demonstrate that simply enforcing structural symmetry with streamlined additive fusion unleashes the network's intrinsic potential.Our baseline, SymUNet, suggests that a well-aligned backbone, even without complex auxiliary modules, is able to handle multiple restoration tasks in a reasonably robust manner across most practical settings, indicating that the reliance on heavy auxiliary modules in prior works may partly stem from compensating for a sub-optimal asymmetric backbone.

Nevertheless, in highly heterogeneous scenarios (e.g., five-task settings), a pixel-based backbone still faces capacity limitations. To break this performance ceiling while maintaining simplicity, we propose SE-SymUNet, which incorporates frozen CLIP features as a semantic anchor to explicitly enhance degradation recognition. By leveraging high-level semantic priors to guide restoration, our model disambiguates complex tasks that are otherwise indistinguishable at the low-level feature stage.

\begin{figure*}[!t]
    \centering
    \setlength{\abovecaptionskip}{2pt} 
    \setlength{\belowcaptionskip}{0pt}
    \includegraphics[width=\textwidth]{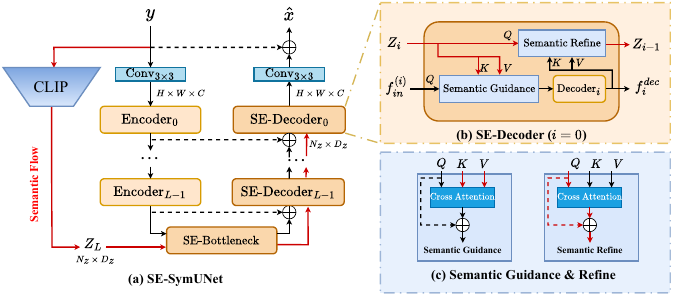}
    \caption{An overview of our semantic enhanced baseline SE-SymUNet, built upon SymUNet. SE-SymUNet introduces a bidirectional semantic guidance module to incorporate CLIP priors. At each decoder stage, image features $\bm f$ are refined by the semantic context $\bm Z$ via the Semantic Guidance Module, followed by updating $\bm Z$ using the refined features $\bm f'$ through the Semantic Refinement Module, forming an adaptive feedback loop. SE-Bottleneck shares the same structure as SE-Decoder.}
    \label{fig:se architecture}
\end{figure*}

\subsection{Primary Baseline: SymUNet}
\vspace{-10pt}

In contrast to recent trends towards complex and asymmetric architectures, our baseline, SymUNet, deliberately returns to a simple, symmetric, and efficient U-shape design. We posit that this streamlined structure is inherently more robust for the all-in-one restoration task.

Our baseline is built on a U-shape architecture with $L$ scales, comprising $L$ encoder layers, $L$ decoder layers, and a bottleneck layer, as illustrated in \cref{fig:architecture}(b). The network begins with an initial convolutional layer that maps the input degraded image $\bm y \in \mathbb{R}^{H \times W \times C_{in}}$ to a higher-dimensional feature space $\bm{f}^{enc}_{0} \in \mathbb{R}^{H \times W \times C}$. Each layer of the encoder, decoder, and bottleneck consists of multiple feature extraction blocks, which share the same structure and are adapted from the efficient Transformer block design in Restormer~\cite{zamir2022restormer}. The network concludes with a final convolutional layer that converts features from the last decoder layer into a residual image, which is added to the input to obtain the final restored image $\bm{\hat{x}} \in \mathbb{R}^{H \times W \times C_{in}}$.

The overall pipeline is as follows:
\begin{equation}
\label{eq:baseline_pipeline_decomposed}
\begin{aligned}
    \bm{f}^{enc}_{0} &= \mathtt{Conv}_{3\times 3}(\bm y), \\
    \bm{s}_{i} &= \mathtt{Encoder}_i(\bm{f}^{enc}_{i}), \quad i \in \{0, \dots, L-1\}, \\
    \bm{f}^{enc}_{i+1} &= \mathtt{DOWN}_{i+1}(\bm{s}_{i}), \quad i \in \{0, \dots, L-1\}, \\
    \bm{f}^{dec}_{L} &= \mathtt{Bottleneck}(\bm{f}^{enc}_{L}), \\
    \bm{f}_{in}^{(i)} &= \mathtt{UP}_{i+1}(\bm{f}^{dec}_{i+1}) + \bm{s}_{i}, \quad i \in \{L-1, \dots, 0\}, \\
    \bm{f}^{dec}_{i} &= \mathtt{Decoder}_i(\bm{f}_{in}^{(i)}), \quad i \in \{L-1, \dots, 0\}, \\
    \bm{\hat{x}} &= \mathtt{Conv}_{3\times 3}(\bm{f}^{dec}_{0}) + \bm y,
\end{aligned}
\end{equation}
where $\mathtt{UP}$ and $\mathtt{DOWN}$ represent upsampling and downsampling operations.

This structure ensures that the skip connection $\bm{s}_{i-1}$ provides the features from the $i$-th encoder stage before they are downsampled, preserving high-resolution details for the corresponding decoder stage.

\subsubsection{Symmetric and Efficient Architecture.}
Our baseline achieves strong performance due to three key differences from \cite{zamir2022restormer,potlapalli2023promptir,tian2025degradation}: 
(\emph{\textbf{i}})~\emph{Symmetric Structure:} the encoder and decoder use equal numbers of blocks for balanced design; 
(\emph{\textbf{ii}})~\emph{Consistent Channel Dimensions:} the decoder maintains a fixed, narrow channel width to encourage compact representations; 
(\emph{\textbf{iii}})~\emph{No Auxiliary Refinement Blocks:} outputs are taken directly from the final decoder stage for simplicity. 
These choices provide a stronger inductive bias for all-in-one restoration.

\subsection{Semantic Enhanced Baseline: SE-SymUNet}

Our SE-SymUNet enhances the primary SymUNet baseline by integrating our bidirectional semantic guidance mechanism, as shown in \cref{fig:se architecture}. This is achieved by effectively wrapping the standard $\mathtt{Bottleneck}$ and $\mathtt{Decoder}$ modules to create their semantic-enhanced counterparts, which we term the $\mathtt{SE\text{-}Bottleneck}$ and $\mathtt{SE\text{-}Decoder}$. The feature extraction blocks in $\mathtt{SE\text{-}Bottleneck}$ and $\mathtt{SE\text{-}Decoder}$ also share the same structure. 
These new composite modules create an adaptive feedback loop between image features and high-level semantic context.

\subsubsection{Bidirectional Semantic Guidance Module.}
The core of our enhancement is a module designed to foster a synergistic dialogue between an image feature map $\bm{f}$ and a semantic context $\bm{Z}$. It consists of two complementary operations:

First, the \textbf{Semantic Guidance} step infuses semantic priors into the restoration pathway. It allows image features to query the semantic context, enabling each local region to draw upon relevant high-level concepts to guide its reconstruction. The operation is defined as:
\begin{equation}
    \mathtt{SemanticGuidance}(\bm{f}, \bm{Z}) = \bm{f} + \mathtt{CA}(\bm{f}, \bm{Z}, \bm{Z}),
\end{equation}
where $\mathtt{CA}$ denotes multi-head cross-attention. To perform this operation, the spatial feature map $\bm f$ is treated as a sequence of patch-based tokens.

Second, the \textbf{Semantic Refinement} step updates the semantic context itself, grounding it in the newly clarified visual evidence. The semantic context queries the refined image features to recalibrate its understanding, making it more accurate for subsequent stages. This operation is defined as:
\begin{equation}
    \mathtt{SemanticRefine}(\bm{f}, \bm{Z}) = \bm{Z} + \mathtt{CA}(\bm{Z}, \bm{f}, \bm{f}).
\end{equation}

\subsubsection{Semantic Context Extraction.}
The semantic context $\bm{Z}$ used in the guidance module is initialized from the input image. We extract an initial context, $\bm{Z}_L$, using the frozen Vision Transformer from CLIP (ViT-L/14). By leveraging the \texttt{last\_hidden\_state}, we obtain a sequence of patch-based tokens, $\bm{Z}_L \in \mathbb{R}^{N_z \times D_z}$ (where $N_z=257$ and $D_z=1024$), which retains granular, spatially-aware information. This $\bm{Z}_L$ serves as the initial state for the semantic context that is then iteratively refined.

\subsubsection{Architectural Integration.}
With the guidance functions defined, we now detail their integration. The encoder architecture remains identical to \textbf{SymUNet}. The upward path, from the bottleneck (level $L$) to the final output (level $0$), is defined by the following iterative process.

Starting with the initial context $\bm{Z}_L$, for each level $i$ from $L$ down to $0$:
\begin{align}
    \bm{f}_{in}^{(i)} &= 
    \begin{cases}
      \bm{f}^{enc}_{L}, & \text{if } i=L \\
      \mathtt{UP}_{i+1}(\bm{f}^{dec}_{i+1}) + \bm{s}_{i}, & \text{if } i < L 
    \end{cases} \\
    \bm{f}'_{in} &= \mathtt{SemanticGuidance}(\bm{f}_{in}^{(i)}, \bm{Z}_{i}), 
\end{align}

\begin{align}
    \bm{f}^{dec}_{i} &= 
    \begin{cases}
      \mathtt{Bottleneck}(\bm{f}'_{in}), & \text{if } i=L \\
      \mathtt{Decoder}_i(\bm{f}'_{in}), & \text{if } i < L 
    \end{cases} \\
    \bm{Z}_{i-1} &= \mathtt{SemanticRefine}(\bm{f}^{dec}_{i}, \bm{Z}_{i}).
\end{align}
This iterative cycle ensures that as the image features become cleaner, the semantic guidance in the next stage becomes correspondingly more precise. The final output is generated identically to the baseline $\bm{\hat{x}} = \mathtt{Conv}_{3\times 3}(\bm{f}^{dec}_{0}) + \bm y$.
\section{Experiments}
\label{sec:exp}

\subsection{Datasets}

We evaluate our models on two multi-task restoration settings: a three-task and a five-task scenario.

\subsubsection{Three-Task All-in-One Restoration.}
A single model is trained for denoising, dehazing, and deraining. The training set combines BSD400~\cite{arbelaez2010contour}, WED~\cite{ma2016waterloo}, RESIDE-$\beta$-OTS~\cite{li2018benchmarking}, and Rain100L~\cite{yang2017deep}, totaling 77,479 images. Evaluation uses CBSD68~\cite{arbelaez2010contour}, SOTS-Outdoor~\cite{li2018benchmarking}, and Rain100L~\cite{yang2017deep} test sets.

\subsubsection{Five-Task All-in-One Restoration.}
We extend to five tasks by adding deblurring and low-light enhancement. The training set is augmented with GoPro~\cite{nah2017deep} and LOL~\cite{Chen2018Retinex}, reaching 80,067 images. Evaluation follows the same benchmarks, with additional GoPro~\cite{nah2017deep} and LOL~\cite{Chen2018Retinex} test sets.

\begin{table}[t]
\renewcommand{\arraystretch}{1.3}
\caption{Quantitative comparison for the three-task all-in-one restoration benchmark. We report PSNR (dB) / SSIM. Best and second-best results are highlighted in \textbf{bold} and \underline{underline}, respectively. \textsuperscript{*} Results are cited from \cite{jiang2025survey}, as these methods were not originally designed for the all-in-one restoration task.}
\label{tab:three_task_results}
\resizebox{\textwidth}{!}{%
\begin{tabular}{|l|c|c|c|c|c|c|}
\hline
\textbf{Method} & \textbf{Dehazing} & \textbf{Deraining} & \multicolumn{3}{c|}{\textbf{Denoising on BSD68}} & \textbf{Average} \\
\hline
 & SOTS-Outdoor & Rain100L & $\sigma=15$ & $\sigma=25$ & $\sigma=50$ & \\
\hline
Restormer\textsuperscript{*}\cite{zamir2022restormer} & 27.78/0.958 & 33.78/0.958 & 33.72/0.865 & 30.67/0.865 & 27.63/0.792 & 30.75/0.901 \\
\hline
NAFNet\textsuperscript{*}\cite{chen2022simple} & 24.11/0.960 & 33.64/0.956 & 33.18/0.918 & 30.47/0.865 & 27.12/0.754 & 29.67/0.844 \\
\hline
AirNet\cite{li2022all} & 27.94/0.962 & 34.90/0.968 & 33.92/0.933 & 31.26/0.888 & 28.00/0.797 & 31.20/0.910 \\
\hline
PromptIR\cite{potlapalli2023promptir} & 30.58/0.974 & 36.37/0.972 & 33.98/0.933 & 31.31/0.888 & 28.06/0.799 & 32.06/0.913 \\
\hline
InstructIR-3D\cite{conde2024instructir} & 30.22/0.959 & 37.98/0.978 & 34.15/0.933 & 31.52/0.890 & 28.30/0.804 & 32.43/0.913 \\
\hline
Perceive-IR\cite{zhang2025perceive} & 30.87/0.975 & 38.29/0.980 & 34.13/0.934 & 31.53/0.890 & 28.31/0.804 & 32.63/0.917 \\
\hline
VLU-Net \cite{zeng2025vision} & 30.71/0.980 & 38.93/0.984 & 34.13/0.935 & 31.48/0.892 & 28.23/0.804 & 32.70/0.919 \\
\hline
MoCE-IR\cite{zamfir2025complexity} & 31.34/0.979 & 38.57/0.984 & 34.11/0.932 & 31.45/0.888 & 28.18/0.800 & 32.73/0.917 \\
\hline
DFPIR\cite{tian2025degradation} & \underline{31.87}/0.980 & 38.65/0.982 & 34.14/0.935 & 31.47/0.893 & 28.25/0.806 & 32.88/0.919 \\
\hline
\textbf{SymUNet (Ours)} & 31.40/\underline{0.981} & \underline{39.12}/\underline{0.985} & \underline{34.22}/\underline{0.937} & \underline{31.57}/\underline{0.894} & \underline{28.32}/\underline{0.808} & \underline{32.93}/\underline{0.921} \\
\hline
\textbf{SE-SymUNet (Ours)} & \textbf{32.02}/\textbf{0.983} & \textbf{39.23}/\textbf{0.986} & \textbf{34.23}/\textbf{0.937} & \textbf{31.58}/\textbf{0.895} & \textbf{28.33}/\textbf{0.809} & \textbf{33.08}/\textbf{0.922} \\
\hline
\end{tabular}
}
\end{table}

\subsection{Implementation Details}
\label{sec:implementation_details}

Both models are built on a 4-level U-shaped architecture ($L=3$, $C=48$) where the encoder, bottleneck, and symmetric decoder comprise [4, 6, 6], [8], and [6, 6, 4] Transformer blocks (MDTA+GDFN), respectively. For SE-SymUNet, semantic contexts from a frozen CLIP (ViT-L/14) are aligned using patch sizes of $p=2$ (bottleneck) and $p=4$ (decoder) for precise injection. We implement models in PyTorch and train on NVIDIA RTX 5880 Ada GPUs using AdamW ($\beta_1=0.9, \beta_2=0.999$) with a cosine schedule ($lr: 10^{-3} \to 10^{-7}$) for 300k (3-task) or 400k (5-task) iterations. The batch size is 32 with $128 \times 128$ crops. To ensure spectral consistency, we define the frequency loss as $\mathcal{L}_{\text{fft}} = \lVert \mathcal{F}(\bm{\hat{x}}) - \mathcal{F}(\bm x) \rVert_1$ and the total objective as $\mathcal{L}_{\text{total}} = \mathcal{L}_1 + \lambda \mathcal{L}_{\text{fft}}$, with the balancing weight $\lambda$ set to 0.1.

\subsection{Main Results}

\begin{table}[t]
\renewcommand{\arraystretch}{1.3}
\caption{Quantitative comparison for the five-task all-in-one restoration benchmark. We report PSNR (dB) / SSIM. Best and second-best results are highlighted in \textbf{bold} and \underline{underline}, respectively. \textsuperscript{*} Results are cited from the survey paper \cite{jiang2025survey}, as the original papers for these methods do not report on this specific all-in-one benchmark.}
\label{tab:five_task_results}
\centering
\resizebox{\linewidth}{!}{%
\begin{tabular}{|l|c|c|c|c|c|c|}
\hline
\textbf{Method} & \textbf{Dehazing} & \textbf{Deraining} & \textbf{Denoising} & \textbf{Deblurring} & \textbf{Low-Light} & \textbf{Average} \\
\hline
 & SOTS-Outdoor & Rain100L & BSD68 ($\sigma=25$) & GoPro & LOL & \\
\hline
Restormer* & 24.09/0.927 & 34.81/0.960 & 31.49/0.884 & 27.22/0.829 & 20.41/0.806 & 27.60/0.881 \\
\hline
NAFNet* & 25.23/0.939 & 35.56/0.967 & 31.02/0.883 & 26.53/0.808 & 20.49/0.809 & 27.76/0.881 \\
\hline
AirNet* & 21.04/0.884 & 32.98/0.951 & 30.91/0.882 & 24.35/0.781 & 18.18/0.735 & 25.49/0.846 \\
\hline
PromptIR* & 26.54/0.949 & 36.37/0.970 & 31.47/0.886 & 28.71/0.881 & 22.68/0.832 & 29.15/0.904 \\
\hline
InstructIR-5D & 27.10/0.956 & 36.84/0.973 & 31.40/0.887 & 29.40/0.886 & 23.00/0.836 & 29.55/0.907 \\
\hline
Perceive-IR & 28.19/0.964 & 37.25/0.977 & \underline{31.44}/0.887 & \underline{29.46}/\underline{0.886} & 22.88/0.833 & 29.84/0.909 \\
\hline
VLU-Net & 30.84/\underline{0.980} & \textbf{38.54}/\underline{0.982} & 31.43/0.891 & 27.46/0.840 & 22.29/0.833 & 30.11/0.905 \\
\hline
MoCE-IR & 30.48/0.974 & 38.04/0.982 & 31.34/0.887 & \textbf{30.05}/\textbf{0.899} & 23.00/0.852 & 30.58/\textbf{0.919} \\
\hline
DFPIR & \underline{31.64}/0.979 & 37.62/0.978 & 31.29/0.889 & 28.82/0.873 & \textbf{23.82}/0.843 & \underline{30.64}/0.913 \\
\hline
\textbf{SymUNet (Ours)} & 31.31/0.979 & 38.05/0.981 & 31.38/\underline{0.891} & 28.12/0.855 & 23.27/\underline{0.858} & 30.43/0.913 \\
\hline
\textbf{SE-SymUNet (Ours)} & \textbf{32.15}/\textbf{0.982} & \underline{38.44}/\textbf{0.983} & \textbf{31.45}/\textbf{0.892} & 28.40/0.864 & 23.22/\textbf{0.861} & \textbf{30.73}/\underline{0.916} \\
\hline
\end{tabular}
}
\end{table}
We present quantitative comparisons against state-of-the-art all-in-one image restoration methods in Tables~\ref{tab:three_task_results} and Table~\ref{tab:five_task_results}. Evaluation on both benchmarks demonstrates the effectiveness of our models, with our symmetric U-Net design establishing a new performance baseline.

\subsubsection{Three-Task All-in-One Restoration}

\begin{figure}[!t]
    \centering
    \includegraphics[width=\textwidth]{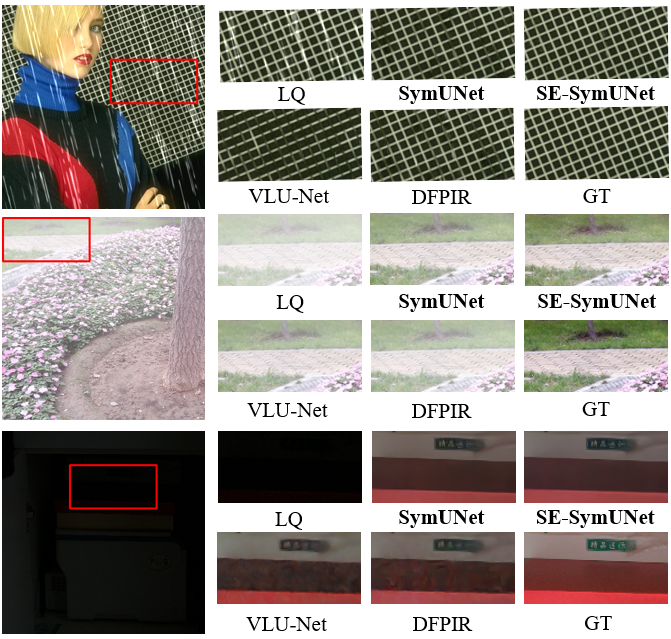}
    \caption{Visual comparison with state-of-the-art methods on the five-task all-in-one benchmark. From top to bottom: deraining, dehazing, and low-light enhancement. SE-SymUNet better removes complex degradations while preserving fine details, restores more natural color and contrast, and recovers details with fewer artifacts under extreme low-light conditions, demonstrating superior visual quality and robustness.}
    \label{fig:visual_comparison_five_task}
\end{figure}
\vspace{-15pt}

As shown in Table~\ref{tab:three_task_results}, our baseline SymUNet achieves an average PSNR of 32.93 dB, surpassing all compared methods and reaching state-of-the-art performance. Notably, as a standard symmetric U-Net without specialized modules, SymUNet demonstrates that a well-balanced architecture with aligned encoder-decoder pathways can effectively handle multi-degradation tasks. Our semantically-enhanced SE-SymUNet further raises the average performance to 33.08 dB, with the largest gain observed on dehazing (+0.62 dB), validating that explicit semantic guidance is particularly beneficial for spatially extensive degradations.

\subsubsection{Five-Task All-in-One Restoration}

As shown in Table~\ref{tab:five_task_results}, on the more challenging scenario with deblurring and low-light enhancement added, SE-SymUNet again achieves state-of-the-art performance with an average PSNR of 30.73 dB, demonstrating strong generalization to a wider range of degradations. The baseline SymUNet also attains a competitive 30.43 dB, underscoring the robustness of our symmetric architecture. SE-SymUNet shows dominant performance on dehazing (+0.51 dB over the second best) and achieves top-tier results on deraining and denoising, indicating that the semantically-enhanced symmetric architecture excels at handling degradations that corrupt fine-grained details. The consistent high performance across five tasks validates our simple symmetric design as an effective foundation for complex multi-degradation restoration, consistent with the qualitative results in \cref{fig:visual_comparison_five_task}. Despite its simplicity, our model is also computationally efficient; detailed comparisons are deferred to the supplementary material.

\subsection{Ablation Study}

We conduct ablation studies on the three-task benchmark with the same settings as Section~\ref{sec:implementation_details}. Results in Table~\ref{tab:ablation_all} quantify the contribution of each component.

\begin{table}[t]
\centering
\caption{Ablation studies on architectural symmetry and semantic guidance.}
\label{tab:ablation_all}
\begin{tabular}{|l|c|c|}
\hline
\textbf{Configuration} & \textbf{PSNR} & \textbf{SSIM} \\
\hline

\multicolumn{3}{|c|}{\textbf{Architectural Symmetry}} \\
\hline
(a) Asymmetric Baseline & 32.55 & 0.919 \\
(b) Symmetric Baseline (SymUNet) & \textbf{32.93} & \textbf{0.921} \\
\hline

\multicolumn{3}{|c|}{\textbf{Semantic Guidance}} \\
\hline
(c) Baseline (SymUNet) & 32.93 & 0.921 \\
(d) + Semantic Guidance (one-way) & 33.02 & 0.921 \\
(e) + Bidirectional Guidance (SE-SymUNet) & \textbf{33.08} & \textbf{0.922} \\
\hline

\end{tabular}
\end{table}

\subsubsection{Impact of Architectural Symmetry}
To verify the effectiveness of our symmetric design, we construct an (a) asymmetric baseline by: 1) doubling the channel dimension after the final skip connection, and 2) appending refinement blocks after the decoder. As shown in Table~\ref{tab:ablation_all}, the symmetric SymUNet achieves 32.93 dB, outperforming (a) by 0.38 dB. This indicates that heavier decoders can hinder efficient information flow, while a symmetric design enables more effective feature fusion.

\subsubsection{Contribution of Semantic Guidance}
We further evaluate the semantic module by progressively adding its components to SymUNet. Introducing the unidirectional `Semantic Guidance' (b) improves PSNR over the baseline, while enabling the bidirectional `Semantic Refine' (c) (i.e., SE-SymUNet) yields the best performance. This validates the advantage of bidirectional interaction and shows that cross-attention-based semantic priors can effectively enhance restoration quality.

\section{Conclusion}

This paper investigates architectural inductive biases in all-in-one image restoration. We demonstrate that structural symmetry—via aligned encoder-decoder hierarchies—outperforms complex asymmetric baselines with lower complexity.Our SymUNet achieves state-of-the-art performance on standard benchmarks, indicating that a streamlined architecture is able to effectively capture diverse degradations in most scenarios. In highly heterogeneous scenarios (e.g., five-task settings) where pixel-based backbones approach their capacity limits, introducing a semantic anchor (SE-SymUNet) further boosts performance. Our findings suggest a pragmatic paradigm: a symmetric backbone ensures efficient feature alignment, while optional semantic injection handles increasing task diversity.


\section*{Acknowledgments}
This work was supported in part by the National Natural Science Foundation of China (62576241, 62172127).

\bibliographystyle{splncs04}
\bibliography{reference}


\appendix

\section{Performance and Efficiency Analysis}
\label{supp:efficiency}

\begin{figure}[!t]
    \centering
    \includegraphics[width=\columnwidth]{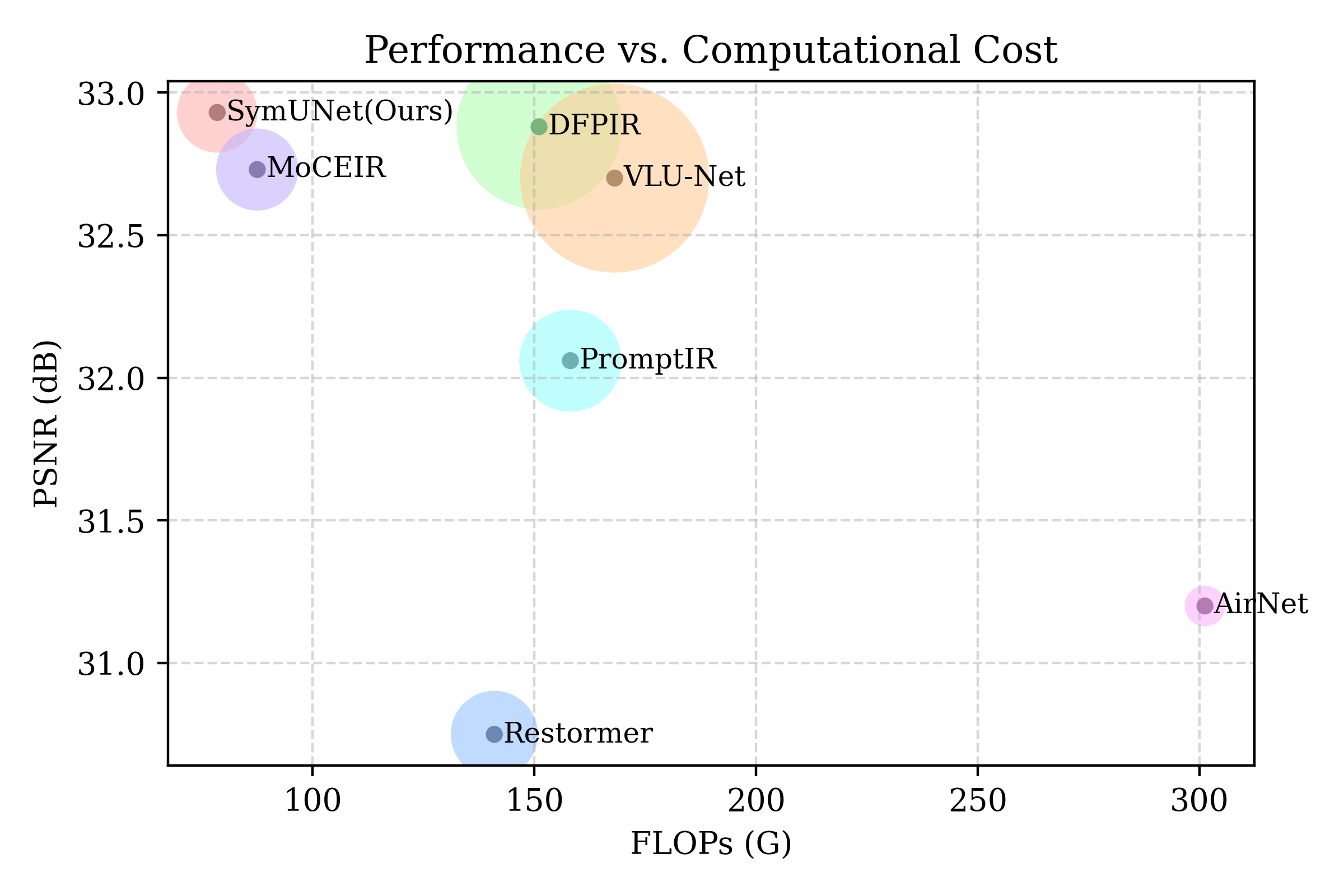}
    \caption{PSNR vs. FLOPs on the three-task benchmark. 
    The bubble size corresponds to the total number of parameters. 
    Our SymUNet has fewer parameters than all competing methods, with the sole exception of AirNet. In terms of performance, SymUNet resides in the optimal top-left quadrant (i.e., representing an optimal efficiency-performance balance), thus achieving state-of-the-art results while featuring substantially lower computational overhead.
    }
    \label{fig:flops_vs_psnr}
\end{figure}

To provide a comprehensive evaluation, we analyze the trade-off between model performance and computational complexity.As illustrated in Figure~\ref{fig:flops_vs_psnr} and detailed in Table~\ref{tab:complexity_comparison}, our proposed methods offer two distinct and compelling design points on the performance-efficiency spectrum.

\begin{table}[h]
\caption{Comparison of performance vs. complexity on the three-task benchmark. Our \textbf{SymUNet} offers unparalleled efficiency, while our \textbf{SE-SymUNet} achieves the highest performance with a highly competitive FLOP count.}
\label{tab:complexity_comparison}

\centering
\begin{tabular}{|l|c|c|c|}
\hline
\textbf{Method} & \textbf{Params (M)} & \textbf{FLOPs (G)} & \textbf{PSNR (dB)} \\
\hline
Restormer~\cite{zamir2022restormer} & 26.10 & 140.99 & 30.75 \\
NAFNet~\cite{chen2022simple} & 17.06 & 15.97 & 29.67 \\
AirNet~\cite{li2022all} & 5.68 & 301.27 & 31.20 \\
PromptIR~\cite{potlapalli2023promptir} & 32.97 & 158.14 & 32.06 \\
Perceive-IR~\cite{zhang2025perceive} & - & - & 32.63 \\
VLU-Net~\cite{zeng2025vision} & 123.00 & 171.07 & 32.70 \\
MoCE-IR~\cite{zamfir2025complexity} & 23.26 & 87.51 & 32.73 \\
DFPIR~\cite{tian2025degradation} & 94.00 & 153.01 & 32.88 \\
\hline
\textbf{SymUNet (Ours)} & 22.26 & 78.47 & 32.93 \\
\textbf{SE-SymUNet (Ours)} & 162.27 & 85.42 & 33.08 \\
\hline
\end{tabular}
\end{table}

Our baseline model, \textbf{SymUNet}, stands out with its exceptional efficiency. With only \textbf{22.26M parameters} and \textbf{78.47G FLOPs}, it achieves a PSNR of \textbf{32.93 dB}, establishing a new state-of-the-art baseline. Compared to DFPIR (94.00M params, 153.01G FLOPs), it is about \textbf{4.2$\times$ smaller} and \textbf{2$\times$ faster} while delivering better performance, demonstrating that a simple symmetric architecture can be both effective and resource-efficient.

Building upon this highly efficient foundation, our enhanced model, \textbf{SE-SymUNet}, focuses on pushing the performance boundary to its peak. It achieves the overall best PSNR of \textbf{33.08 dB}, setting a new state-of-the-art for all-in-one restoration. This superior result is attributed to our novel bidirectional semantic guidance module, which effectively leverages high-level priors to resolve complex degradations. As shown in Table~\ref{tab:complexity_comparison}, this performance leap is attained with a remarkably small increase in computational load (FLOPs), confirming that our guidance mechanism is computationally lean and does not create an inference bottleneck. This approach—augmenting a strong, efficient baseline with targeted semantic knowledge—presents a more effective strategy than relying on computationally expensive backbones. Together, our models offer a clear choice: SymUNet for an optimal efficiency-performance balance, and SE-SymUNet for scenarios where achieving the absolute highest restoration quality is the primary goal.

\section{Feature Visualization Supporting Structural Symmetry}
\label{supp:feature_visualization}
\begin{figure}[t]
    \centering
    \includegraphics[width=\columnwidth]{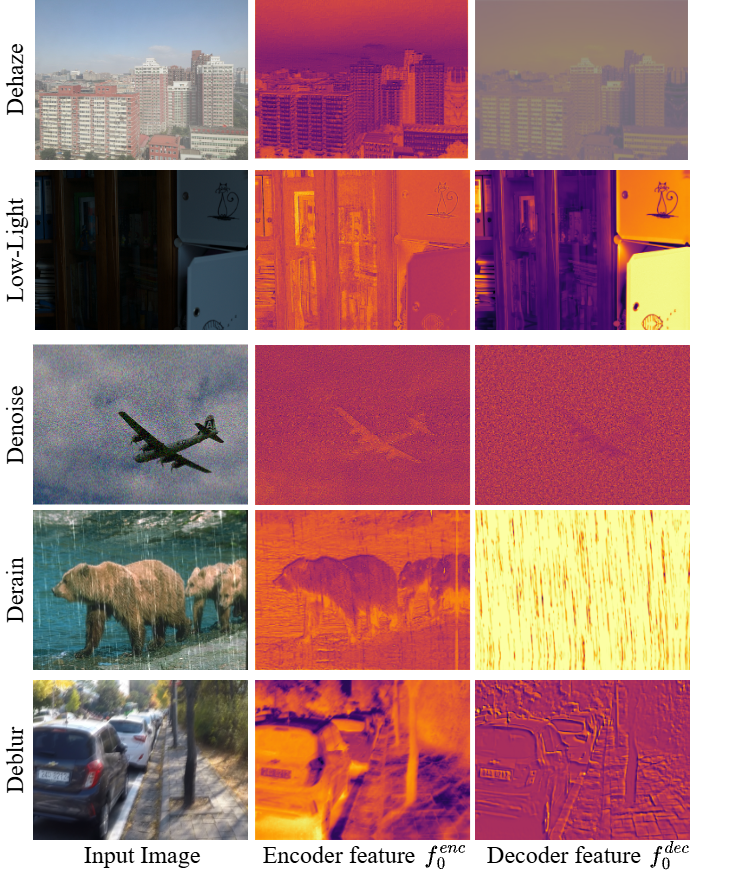}
    \caption{Visualization of degradation-carrying features learned by our SymUNet.}
    \label{fig:residual_visualization}
\end{figure}

To further validate our core motivation, we present a visualization of the degradation-aware features learned by SymUNet in Figure~\ref{fig:residual_visualization}. 
As shown in the figure, the encoder feature $\bm{f}_0^{enc}$ inherently encodes degradation information. More importantly, these features are consistently preserved and propagated through symmetric skip connections, resulting in decoder features that can explicitly disentangle different degradation patterns.
This observation further supports our claim that, when properly designed, architectural simplicity itself can serve as a strong inductive bias for all-in-one image restoration.

\section{Visual Results}
\label{supp:visual}

\begin{figure}
    \centering
    \includegraphics[width=\textwidth]{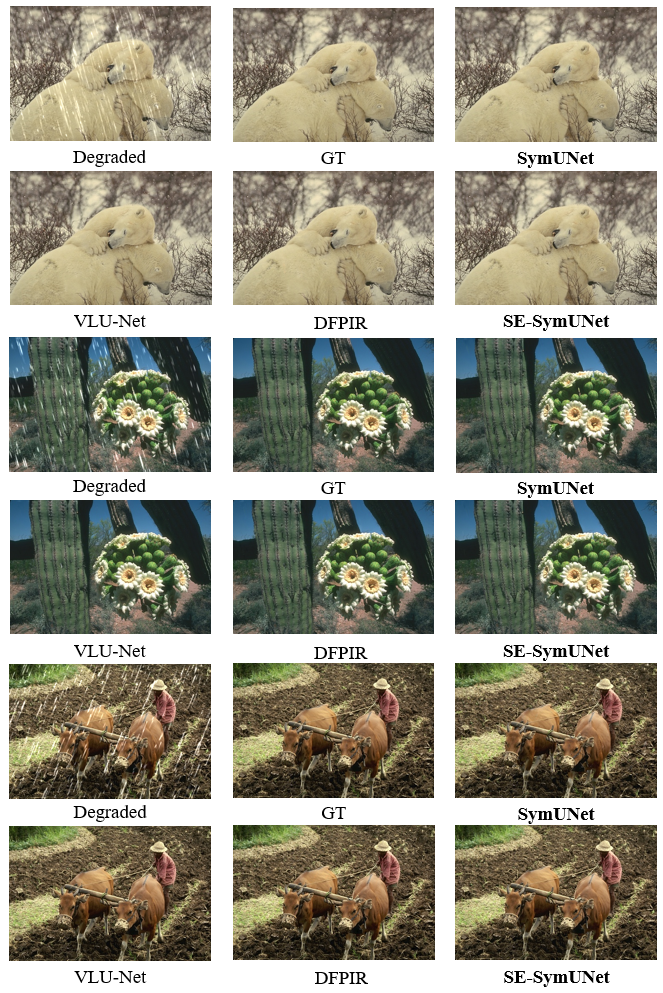}
    \caption{Visual results on Rain100L dataset.}
    \label{fig:derain_visualization}
\end{figure}

\begin{figure}[t]
    \centering
    \includegraphics[width=\textwidth]{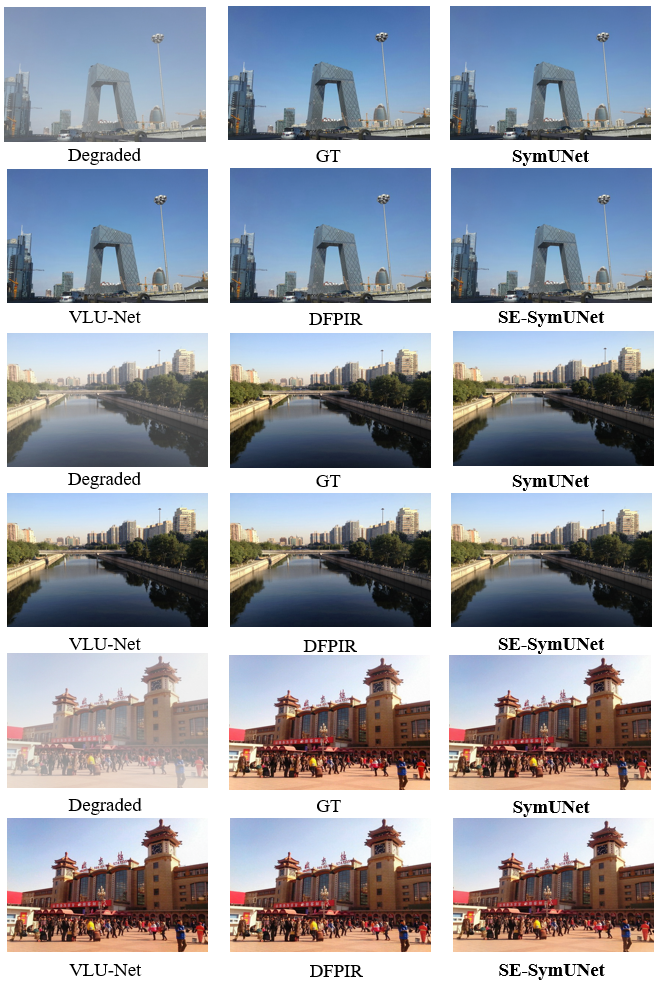}
    \caption{Visual results on SOTS dataset.}
    \label{fig:dehaze_visualization}
\end{figure}

\begin{figure}[t]
    \centering
    \includegraphics[width=\textwidth]{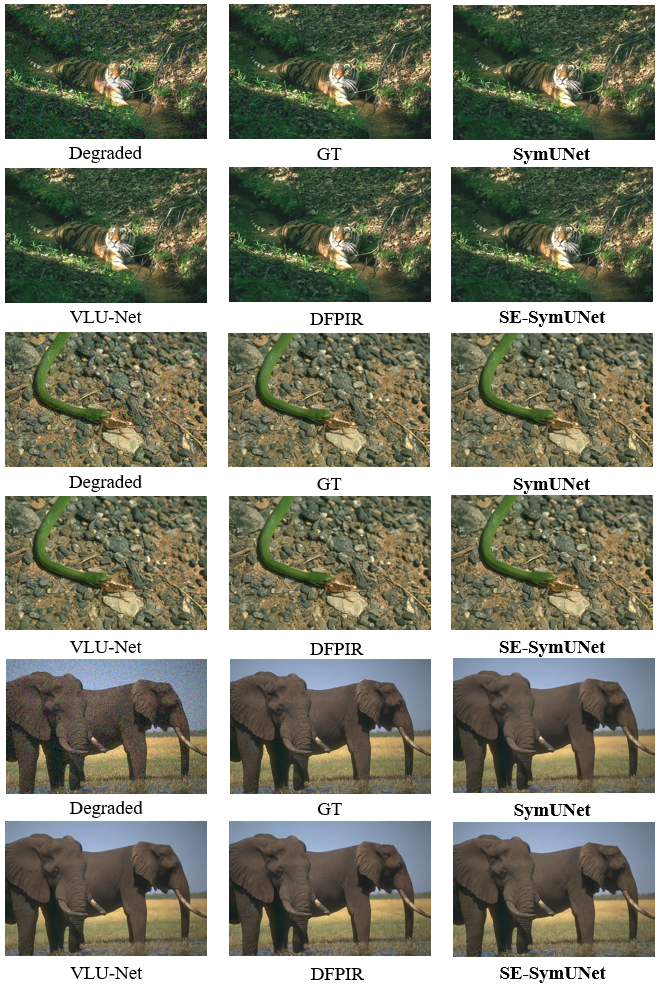}
    \caption{Visual results on CBSD68 dataset.}
    \label{fig:denoise_visualization}
\end{figure}

\begin{figure}[t]
    \centering
    \includegraphics[width=\textwidth]{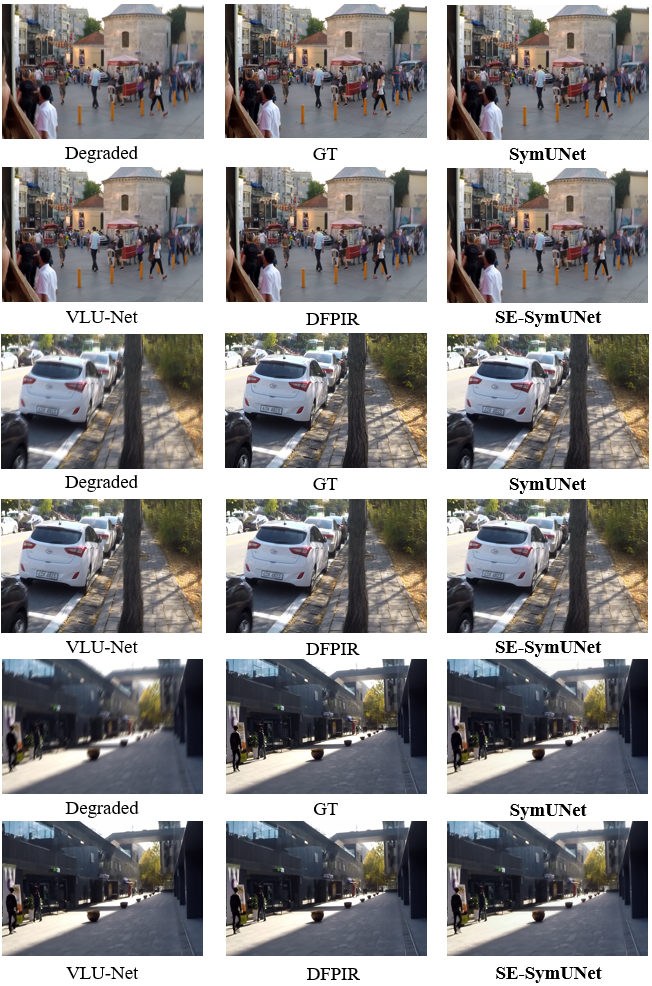}
    \caption{Visual results on GoPro dataset.}
    \label{fig:deblur_visualization}
\end{figure}

\begin{figure}[t]
    \centering
    \includegraphics[width=\textwidth]{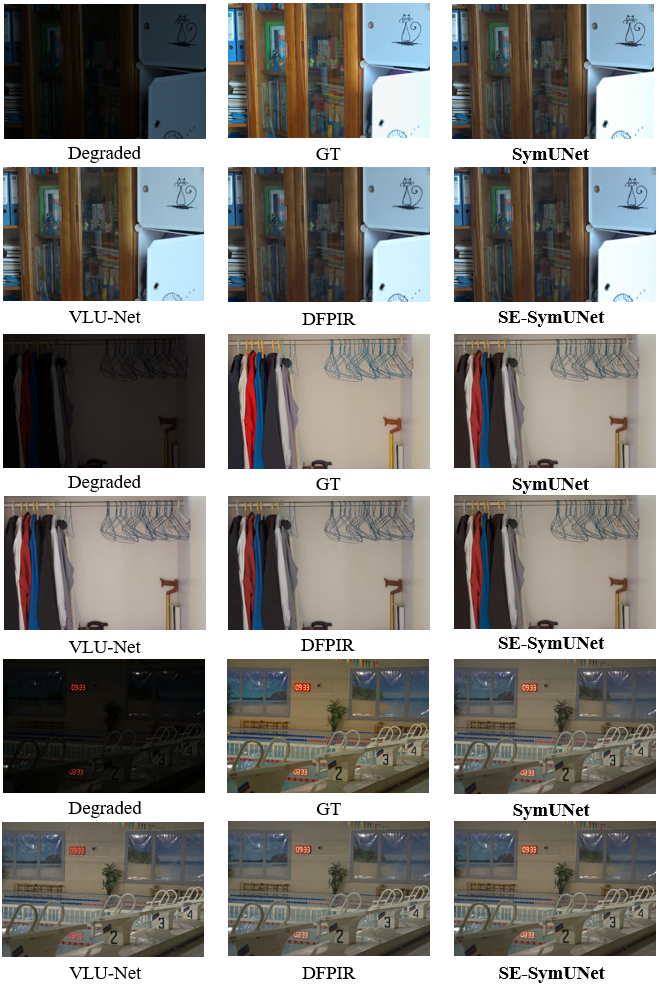}
    \caption{Visual results on LOL dataset.}
    \label{fig:low_light_visualization}
\end{figure}

We provide visual comparisons across different image restoration tasks: image deraining results are shown in \cref{fig:derain_visualization}; image dehazing results are shown in \cref{fig:dehaze_visualization}; image denoising results are shown in \cref{fig:denoise_visualization}; image deblurring results are shown in \cref{fig:deblur_visualization}; image low-light enhancement results are shown in \cref{fig:low_light_visualization}.

\vspace{-15pt}

\end{document}